\documentclass{article} 
\usepackage{iclr2026_conference,times}

\usepackage{amsmath,amsfonts,bm}

\def\eqref#1{equation~\ref{#1}}

\def\1{\bm{1}}

\DeclareMathAlphabet{\mathsfit}{\encodingdefault}{\sfdefault}{m}{sl}
\SetMathAlphabet{\mathsfit}{bold}{\encodingdefault}{\sfdefault}{bx}{n}

\usepackage{hyperref}
\usepackage{url}
\usepackage{graphicx}
\usepackage{algorithm}
\usepackage{algorithmicx}
\usepackage{algpseudocode}
\usepackage{booktabs}
\title{TIE: A Training–Inversion–Exclusion \\Framework for Visually Interpretable \\and Uncertainty-Guided \\Out-of-Distribution Detection}

\author{P. Suhail, R. Afroz, A Sethi \\
Department of Electrical Engineering\\
IIT Bombay\\
\texttt{\{psuhail,rafroz,asethi\}@iitb.ac.in} \\
}

\iclrfinalcopy 
\begin{document}

\maketitle

\begin{abstract}
Deep neural networks often struggle to recognize when an input lies outside their training experience, leading to unreliable and overconfident predictions. Building dependable machine learning systems therefore requires methods that can both estimate predictive \textit{uncertainty} and detect \textit{out-of-distribution (OOD)} samples in a unified manner. In this paper, we propose \textbf{TIE: a Training--Inversion--Exclusion} framework for visually interpretable and uncertainty-guided anomaly detection that jointly addresses these challenges through iterative refinement. TIE extends a standard $n$-class classifier to an $(n+1)$-class model by introducing a garbage class initialized with Gaussian noise to represent outlier inputs. Within each epoch, TIE performs a closed-loop process of \textit{training, inversion, and exclusion}, where highly uncertain inverted samples reconstructed from the just-trained classifier are excluded into the garbage class. Over successive iterations, the inverted samples transition from noisy artifacts into visually coherent class prototypes, providing transparent insight into how the model organizes its learned manifolds. During inference, TIE rejects OOD inputs by either directly mapping them to the garbage class or producing low-confidence, uncertain misclassifications within the in-distribution classes that are easily separable, all without relying on external OOD datasets.  A comprehensive threshold-based evaluation using multiple OOD metrics and performance measures such as \textit{AUROC}, \textit{AUPR}, and \textit{FPR@95\%TPR} demonstrates that TIE offers a unified and interpretable framework for robust anomaly detection and calibrated uncertainty estimation (UE) achieving near-perfect OOD detection with \textbf{\(\!\approx\!\) 0 FPR@95\%TPR} when trained on MNIST or FashionMNIST and tested against diverse unseen datasets.

\end{abstract}

\section{Introduction}

The rapid integration of machine learning models into safety-critical domains such as autonomous driving, medical diagnosis, and financial decision-making has made \textit{model reliability} and \textit{robustness} indispensable. Despite remarkable progress in predictive accuracy, modern neural networks remain vulnerable to overconfidence, often producing high-confidence predictions even for inputs far removed from the training distribution \cite{suhail2025networkinversiongeneratingconfidently}. Such behavior compromises safety and trustworthiness, motivating the need for models that can not only recognize when they are uncertain but also identify when an input is fundamentally \textit{out-of-distribution (OOD)}. The complementary tasks of \textbf{OOD detection}—discriminating in-distribution from anomalous inputs—and \textbf{uncertainty estimation (UE)}—quantifying predictive confidence—are therefore essential for building dependable, interpretable, and fail-safe learning systems.

Although closely related, OOD detection and UE have traditionally been treated as disjoint problems. Most existing methods rely on post-hoc calibration or require access to auxiliary OOD datasets for fine-tuning, which limits scalability and interpretability. Furthermore, post-hoc uncertainty estimates often fail to capture epistemic uncertainty, as they are derived from fixed, discriminative models without revisiting how the underlying decision boundaries are formed. Consequently, there remains a need for frameworks that integrate uncertainty modeling directly into the learning process while maintaining interpretability and eliminating reliance on external supervision.

In this paper, we introduce \textbf{TIE (Training--Inversion--Exclusion)}, a unified framework that jointly addresses uncertainty estimation and OOD detection through a self-refining training mechanism. Building upon the concept of network inversion \cite{suhail2024networkcnn}, TIE reconstructs input samples from classifier outputs to visualize and analyze class manifolds, thereby linking interpretability with uncertainty modeling. Within each epoch, TIE performs a cyclic process of \textit{training, inversion, and exclusion}, where inverted reconstructions are assessed using a dynamic uncertainty threshold. Samples exhibiting high uncertainty are excluded into the garbage class, enabling the model to progressively purify its decision boundaries and enhance robustness. The framework extends a standard $n$-class classifier to an $(n+1)$-class architecture by introducing a dedicated garbage class initialized with Gaussian noise to represent outlier content. 

Over successive iterations, the inverted samples evolve from noisy artifacts into visually coherent class prototypes, offering interpretable insights into the structure of each class manifold and how uncertainty evolves during learning. During inference, TIE effectively rejects OOD samples by assigning them either directly to the garbage class or to any of the in-distribution class with relatively low confidence, making them easily separable by thresholding. Our results demonstrate that TIE provides a visually interpretable, uncertainty-aware, and self-correcting framework that unifies OOD detection and uncertainty estimation without reliance on post-hoc calibration or external OOD datasets.

\section{Prior Work}
\label{gen_inst}

\paragraph{Network Inversion.}
Inversion aims to reconstruct inputs that elicit desired outputs from a neural network. Early work on multilayer perceptrons employed gradients to do reconstructions, however these were often noisy in appearance~\cite{KINDERMANN1990277,784232,SAAD200778}. Later ~\cite{Wong2017NeuralNI} explored evolutionary optimization and constrained search, while subsequent studies improved visual fidelity through explicit priors such as smoothness constraints or pretrained generative models~\cite{mahendran2015understanding,yosinski2015understanding,mordvintsev2015inceptionism,nguyen2016synthesizing,nguyen2017plug}. The relationship between inversion and adversarial examples became evident, as unconstrained inversion frequently converged to adversarial artifacts~\cite{szegedy2013intriguing,goodfellow2014explaining}. In contrast, adversarially robust networks were shown to yield more human-aligned reconstructions~\cite{tsipras2018robustness,engstrom2019adversarial,santurkar2019image}. Recently surrogate loss learning for stable inversion~\cite{liu2022landscapelearningneuralnetwork}, generative modeling conditioned on classifier outputs~\cite{suhail2024networkcnn}, and logical formulations using CNF  constraints ~\cite{suhail2024network} for deterministic inversion have been explored.

\paragraph{Out-of-Distribution Detection.}
OOD detection focuses on identifying test inputs that lie outside the distribution of the training data. \cite{hendrycks2018baselinedetectingmisclassifiedoutofdistribution} introduced the \textbf{Maximum Softmax Probability (MSP)} score as a simple yet effective measure, showing that correctly classified samples tend to yield higher softmax probabilities compared to misclassified or OOD inputs. Building upon this, \cite{liang2020enhancingreliabilityoutofdistributionimage} proposed \textbf{ODIN}, which enhances separability between in- and out-of-distribution samples by applying temperature scaling and small input perturbations, significantly reducing false positives.
Subsequently, \cite{liu2021energybasedoutofdistributiondetection} reformulated OOD detection through an \textbf{energy-based} framework, demonstrating that energy scores aligned with the input’s log-likelihood offer superior separation between in- and out-of-distribution data while addressing softmax overconfidence. In parallel, \cite{lee2018simpleunifiedframeworkdetecting} proposed the \textbf{Mahalanobis distance}-based approach, which models class-conditional feature distributions via Gaussian discriminant analysis to compute confidence scores, achieving strong performance on both adversarial and natural OOD detection tasks. Recent works like SCOOD~\cite{fanlu2023scood} enhance semantic coherence in OOD detection through uncertainty-aware optimal transport and adaptive cost modeling. Gaussian process-based techniques~\cite{chen2024uncertainty} model predictive uncertainty using only in-distribution data, while normalizing flow–based models such as PostNet~\cite{charpentier2020postnet} learn posteriors over predictive probabilities for reliable OOD discrimination without explicit OOD supervision.

\paragraph{Uncertainty Quantification(UQ).}
UQ has become fundamental to building trustworthy AI systems, especially where overconfident errors can have severe consequences. Post-hoc methods are widely used due to their compatibility with pretrained classifiers. Monte Carlo Dropout (MC Dropout)~\cite{gal2016dropout} introduces stochastic inference to approximate Bayesian model averaging, while temperature scaling~\cite{guo2017calibration} improves calibration via a single scalar parameter. Evidential Deep Learning~\cite{sensoy2018evidential} models classification as a Dirichlet evidence estimation problem, while DEUP~\cite{jain2022deup} predicts generalization error using a learned uncertainty regressor and evidential meta-models~\cite{shen2023posthoc} generate Dirichlet parameters directly from classifier embeddings. Bayesian neural networks (BNNs)~\cite{neal1996bayesian,blundell2015weight} and variational inference approaches estimate posterior weight distributions, whereas Deep Ensembles~\cite{lakshminarayanan2017simple} combine multiple independently trained networks for superior calibration and robustness under shift. Domain-specific extensions include test-time augmentation, uncertainty-aware segmentation~\cite{jungo2020uncertainty}, and Bayesian approximations for volumetric imaging~\cite{kwon2020uncertainty}. BAY-MED~\cite{bala2025baymed} extends evidential meta-modeling to breast cancer classification, achieving improved robustness against OOD samples. Autoinverse~\cite{ansari2022autoinverse} integrates predictive uncertainty into the inversion process, constraining reconstructions toward reliable training regions to improve robustness. 

\paragraph{Our Work.}
Prior research underscores the complementary nature of out-of-distribution detection and uncertainty estimation, yet most methods treat them as loosely coupled or rely on post-hoc calibration. These gaps motivate the need for a unified, self-refining approach— \textbf{TIE}—that integrates uncertainty guidance, inversion-based interpretability, and dynamic exclusion within a single iterative learning process to achieve uncertainty-aware OOD detection.

\section{Preliminaries -- Network Inversion}
\label{sec3}

The cornerstone of the \textbf{TIE} framework is the \textit{inversion} ~\cite{suhail2024networkcnn} process that aims to reconstruct inputs that evoke specific outputs of a neural network. In conventional formulations, inversion is applied \textit{post hoc} on a fully trained classifier to analyze or reconstruct its training distribution. We, in contrast, embed the inversion process \textit{within} the classifier training itself, allowing the inversion dynamics to co-evolve.

This method relies on the input-output relationship of a classifier \( f_\theta: \mathcal{X} \rightarrow \Delta^{K-1} \), where \(\mathcal{X}\) is the input space and \( \Delta^{K-1} \) is the \((K-1)\)-dimensional probability simplex over class labels. Formally, we train a conditional generator \( \mathcal{G}_\phi: \mathcal{Z} \times \mathbb{R}^K \rightarrow \mathcal{X} \), parameterized by \(\phi\), to invert the classifier’s behavior. Instead of conditioning the generator directly on a discrete class label \( y \in \{1, \dots, K\} \), we adopt a soft conditioning strategy based on sampled vectors. Specifically, in addition to the latent input \( z \sim \mathcal{N}(0, I) \), we generate a conditioning vector \( v \in \mathbb{R}^K \), where each component is independently sampled from a standard normal distribution:\(
v_k \sim \mathcal{N}(0, 1), \quad \text{for } k = 1, \dots, K.
\)
The vector \(v\) is then transformed into a probability distribution \( \tilde{y} \in \Delta^{K-1} \) using the softmax function.
The generator thus receives the pair \( (z, \tilde{y}) \) and produces a synthetic input \( \hat{x} = \mathcal{G}_\phi(z, \tilde{y}) \) intended to be classified as \( y = \arg\max_k \tilde{y}_k \).

This formulation allows the generator to be softly conditioned on class identity, without directly exposing the true label enabling multiple conditioning vectors to correspond to the same class, but with different softmax confidence levels. Model inversion aims to recover a plausible input \( x \in \mathcal{X} \) that minimizes a label-consistent objective:\(
\hat{x} = \arg\min_{x \in \mathcal{X}} \mathcal{L}_{\text{inv}}(f_\theta(x), y),
\)
where \( \mathcal{L}_{\text{inv}} \) encourages alignment between the classifier output and the target label, augmented with regularizers for realism or diversity. In our setup, this is realized by optimizing a conditional generator \( \mathcal{G}_\phi(z, y) \) to minimize a composite loss:
\[
\mathcal{L}_{\text{Inv}} = 
\alpha \cdot \mathcal{L}_{\text{KL}} +
\beta \cdot \mathcal{L}_{\text{CE}} +
\gamma \cdot \mathcal{L}_{\text{Cosine}}
\]
\[
\mathcal{L}_{\text{Inv}} = 
\alpha \cdot \sum_{k=1}^K y_k \log \left( \frac{y_k}{f_\theta^k(\hat{x})} \right) + \beta \cdot \left( -\sum_{k=1}^K y_k \log f_\theta^k(\hat{x}) \right) + \gamma \cdot \frac{1}{N(N-1)}
\sum_{i \neq j} \left( 1 - \frac{\langle h_i, h_j \rangle}{\|h_i\| \cdot \|h_j\|} \right)
\]
where \( y \) is the target (soft or one-hot) label distribution, \( f_\theta^k(\hat{x}) \) is the classifier's softmax probability for class \( k \), \( h_i \) and \( h_j \) are feature embeddings from generated samples \( \hat{x}_i \), \( \hat{x}_j \), and \( N \) is the batch size.

The \textbf{Cross-Entropy Loss} \( \mathcal{L}_{\text{CE}} \) aligns generated images with their target classes, while the \textbf{KL Divergence} \( \mathcal{L}_{\text{KL}} \) ensures that the classifier’s output distribution matches the conditioning vector, capturing subtle uncertainty variations. The \textbf{Cosine Similarity Loss} \( \mathcal{L}_{\text{Cosine}} \) enforces feature-level diversity across generated samples, mitigating collapse to overly prototypical representations, while the coefficients \( \alpha, \beta, \gamma \) balance these objectives. This objective encourages the generator to synthesize samples that not only match the classifier’s target output but also exhibit high inter and intra-class diversity that is representative of outliers.

\begin{figure}[t]
\centering
\includegraphics[width=0.95\linewidth]{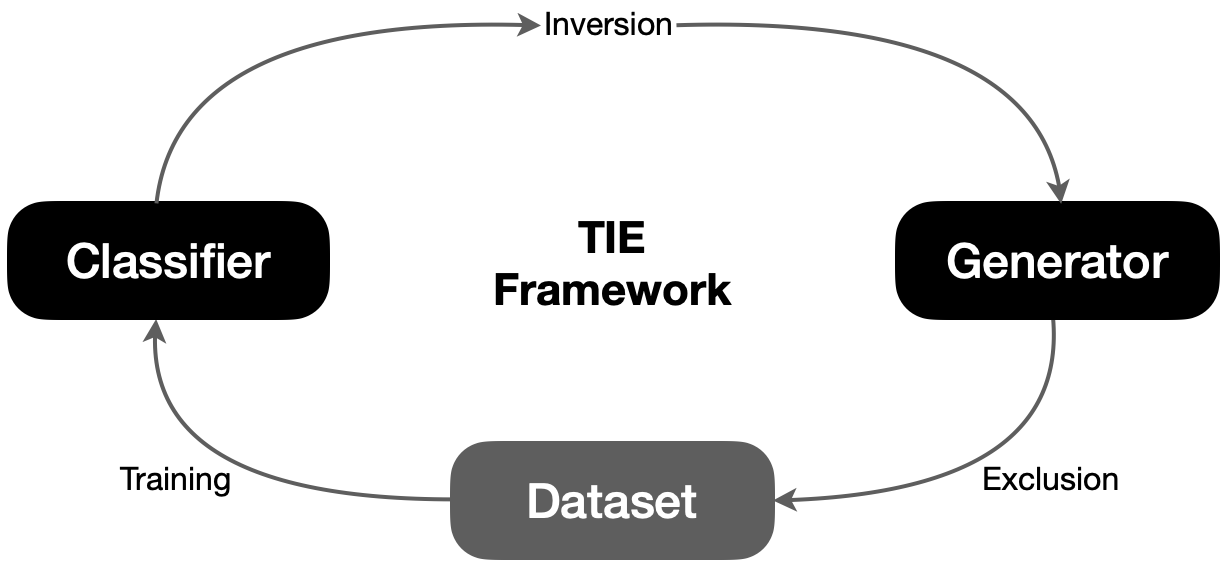}
\caption{
Overview of the proposed \textbf{TIE} framework.
}
\label{fig:tie_framework}
\end{figure}

\begin{algorithm}[t]
\caption{TIE: Training--Inversion--and--Exclusion Framework}
\label{alg:tie}
\begin{algorithmic}[1]
\Require Training data $\mathcal{D}_{\mathrm{train}}=\{(x_i,y_i)\}_{i=1}^N$; number of classes $n$; classifier $f_{\theta}$; generator $\mathcal{G}_{\phi}$; epochs $T$; uncertainty threshold $\lambda$
\Ensure Trained $f_{\theta}$ for unified OOD Detection and UE
\vspace{3pt}

\State \textbf{Learning: TIE Framework}
\State Initialize garbage class $(n{+}1)$ with Gaussian noise $\mathcal{D}_{\mathrm{garbage}}\!\sim\!\mathcal{N}(0,I)$
\State Initialize class weights $\mathbf{w}_0$ for weighted cross-entropy

\For{$t=1$ to $T$} \Comment{--- Epoch-level Training–Inversion–Exclusion Loop ---}
  \State \textbf{Compute class weights:} \\
         $\mathbf{w}_t\!\leftarrow\!\text{proportions in }(\mathcal{D}_{\mathrm{train}}\!\cup\!\mathcal{D}_{\mathrm{garbage}})$
  \State \textbf{Train classifier:} \\
         update $f_{\theta}$ on $\mathcal{D}_{\mathrm{train}}\!\cup\!\mathcal{D}_{\mathrm{garbage}}$ using $\mathcal{L}_{\mathrm{CE}}(\mathbf{w}_t)$
  \State \textbf{Learn generator via inversion:} 
         for each class $c\!\in\!\{1,\dots,n{+}1\}$, sample $z\!\sim\!\mathcal{N}(0,I)$, generate $\hat{x}_c\!=\!\mathcal{G}_{\phi}(z,\tilde{y}_c)$, 
         and optimize $\phi$ using $\mathcal{L}_{\mathrm{Inv}}$ \Comment{Section.~\ref{sec3}}
  \State Collect inverted samples $\mathcal{D}_{\mathrm{inv}}\!=\!\{\hat{x}_c\}_{c=1}^{n+1}$

  \State \textbf{Compute dynamic uncertainty threshold:}
  \For{each $x\!\in\!\mathcal{D}_{\mathrm{train}}$}
      \State $u(x)\!=\!\mathrm{UE}(f_{\theta}(x))$  \Comment{Section.~\ref{eq:uncertainty_estimate}}
  \EndFor
  \State $\mu_t\!=\!\mathrm{mean}(u),\; \sigma_t\!=\!\mathrm{std}(u),\;
         \tau_t\!=\!\mu_t+\lambda\sigma_t$

  \State \textbf{Exclude high-uncertainty inversions:}
  \For{each $\hat{x}\!\in\!\mathcal{D}_{\mathrm{inv}}$}
     \If{$\mathrm{UE}(f_{\theta}(\hat{x}))>\tau_t$}
        \State $\mathcal{D}_{\mathrm{garbage}}\!\leftarrow\!
                \mathcal{D}_{\mathrm{garbage}}\!\cup\!\{\hat{x}\}$  \Comment{send to garbage}
     \EndIf
  \EndFor
  \State Update $\mathbf{w}_{t{+}1}$ for next epoch
\EndFor
\vspace{3pt}

\State \textbf{Inference: Two-Level OOD Detection}
\State For a test sample $x^{\ast}$, compute $\mathbf{p}\!=\!f_{\theta}(x^{\ast})$ and $u\!=\!\mathrm{UE}(\mathbf{p})$
\If{$\arg\max_i \mathbf{p}_i\!=\!n{+}1$}
   \State \textbf{Level-1:} Direct OOD Detection into garbage class.
\ElsIf{$u>\tau_T$}
   \State \textbf{Level-2:} Fine-grained OOD Detection via thresholding.
\Else
   \State Assign in-distribution label $\arg\max_i \mathbf{p}_i$
\EndIf
\end{algorithmic}
\end{algorithm}

\section{The TIE Framework}
\label{sec:tie}

The proposed \textbf{TIE (Training--Inversion--Exclusion)} framework builds up on ~\cite{suhail2025networkinversionuncertaintyawareoutofdistribution} to unify OOD detection and UE into a single, interpretable, and self-refining learning paradigm as summarized in Algorithm \ref{alg:tie}. TIE is designed as an iterative optimization process that tightly couples classifier learning, network inversion, and uncertainty-guided exclusion, enabling the model to progressively refine its decision boundaries and calibrate uncertainty over time.

\subsection{Classifier Extension}
For a standard $n$-class classification task, TIE extends the classifier $f_\theta: \mathcal{X} \rightarrow \Delta^{n}$ into an $(n+1)$-class formulation by introducing an auxiliary garbage class. This additional class acts as a receptacle for anomalous or uncertain samples that do not belong to any of the known in-distribution categories. At initialization, the garbage class is populated with random Gaussian noise, providing a background reference distribution that enables the classifier to learn a preliminary separation between structured and unstructured regions of the input space.

\subsection{Embedded Inversion}
Unlike conventional formulations where inversion is applied on a fully trained classifier, TIE embeds the inversion mechanism directly into the training process. A conditional generator $\mathcal{G}_\phi: \mathcal{Z} \times \mathbb{R}^{n+1} \rightarrow \mathcal{X}$ is co-trained with the classifier $f_\theta$, forming a bi-directional loop where both networks evolve together. After each training epoch, $\mathcal{G}_\phi$ reconstructs representative samples corresponding to all output classes, guided by the inversion objective $\mathcal{L}_{\text{Inv}}$ introduced in Section~\ref{sec3}. These reconstructed samples approximate the classifier’s current perception of each class manifold, reflecting how confidently or coherently the model represents its learned distribution. Early in training, the inverted samples are typically noisy or semantically inconsistent—revealing regions of high uncertainty. As the classifier improves, the generator’s outputs gradually evolve into visually coherent class prototypes, providing transparent visual evidence of how decision boundaries sharpen and stabilize. This co-evolution ensures that inversion not only interprets the classifier but also actively participates in shaping its internal representation space.

\subsection{Iterative Refinement Cycle}
Each epoch in TIE consists of three interconnected stages—\textbf{Training}, \textbf{Inversion}, and \textbf{Exclusion}—that operate in a closed feedback loop as shown in Figure \ref{alg:tie}:
\begin{enumerate}
    \item \textbf{Training:} The classifier $f_\theta$ is trained on the combined dataset containing both in-distribution samples and garbage samples using a weighted cross-entropy objective to correct for class imbalance caused by the addition of inverted samples into the garbage class.
    \item \textbf{Inversion:} The generator $\mathcal{G}_\phi$ reconstructs input samples conditioned on each class’s logits, optimizing the inversion loss $\mathcal{L}_{\text{Inv}}$ to reflect the current state of the classifier’s learned manifold.
    \item \textbf{Exclusion:} The reconstructed samples are evaluated for confidence and coherence using the classifier’s predictive uncertainty. Samples exceeding a dynamic uncertainty threshold $\tau_t$ are assigned to the garbage class.
\end{enumerate} 
This refinement continues over successive epochs until the model learns to separate structured data from anomalous regions, visually through clearer reconstructions and probabilistically through calibrated uncertainty estimates. The continual feedback between inversion and exclusion drives convergence toward robust, well-formed decision boundaries.

\subsection{Uncertainty-Guided Exclusion.}
Uncertainty forms the backbone of the exclusion phase. After each epoch, the predictive uncertainty for all training samples is computed using the classifier’s softmax distribution as
\[
\text{UE}(\mathbf{p}) = 1 - 
\frac{\sum_{i=1}^{n+1} \left(p_i - \frac{1}{n+1}\right)^2}
{\sum_{i=1}^{n+1} \left(\delta_{i,k} - \frac{1}{n+1}\right)^2},
\label{eq:uncertainty_estimate}
\]
where $k = \arg\max_i p_i$ and $\delta_{i,k}$ denotes the Kronecker delta. This normalized deviation from uniformity quantifies the model’s confidence: lower values indicate sharp, confident predictions, while higher values signal uncertainty.

The uncertainty scores of the \emph{training data} are aggregated at each epoch to compute a dynamic exclusion threshold,
\(
\tau_t = \mu_t + \lambda \sigma_t,
\)
where $\mu_t$ and $\sigma_t$ represent the mean and standard deviation of the training uncertainties at epoch $t$, and $\lambda$ is a scaling parameter that controls the strictness of the exclusion. As training progresses, the classifier becomes progressively more confident on in-distribution samples, leading to a natural decline in uncertainty and a correspondingly tighter threshold $\tau_t$. 

This threshold is then applied to the uncertainties of the \emph{inverted samples} to determine which reconstructions are unreliable. Samples with $u_i > \tau_t$ are considered incoherent or anomalous and are assigned to the garbage class for the next training cycle. This evolving threshold ensures that exclusion criteria adapt automatically to the classifier’s calibration, progressively filtering out uncertain or OOD reconstructions while reinforcing confident class boundaries.

\subsection{Inference and Evaluation}

Once training converges, the TIE framework performs joint \textbf{OOD detection} and \textbf{UE} during inference. The inclusion of the garbage class enables TIE to perform two-tier OOD detection for both coarse- and fine-grained anomalies while maintaining high in-distribution performance.
\begin{enumerate}
    \item \textbf{Direct OOD Detection into Garbage Class.}  
    At this first level, the model’s $(n{+}1)$-class formulation allows most OOD inputs to be \textit{directly classified into the garbage class}, enabling coarse anomaly rejection without post-hoc calibration. The goal here is to ensure that the classifier maintains strong predictive accuracy on in-distribution data while simultaneously recognizing clear OOD samples. This stage is evaluated using \textbf{predictive accuracy} over both in- and out-of-distribution sets: the model is trained on one dataset and tested on others treated as OOD sources. 

    \item \textbf{Threshold-Based Fine-Grained OOD Detection.}  
    Some OOD samples lie close to class boundaries and may be misclassified into in-distribution categories with relatively low confidence. To detect such subtle anomalies, TIE employs a threshold-based analysis using multiple \textbf{OOD metrics}—including \textit{UE} [~\ref{eq:uncertainty_estimate}],
    \textit{ODIN}~\cite{liang2020enhancingreliabilityoutofdistributionimage},
    \textit{Energy}~\cite{liu2021energybasedoutofdistributiondetection}, \textit{MSP}~\cite{hendrycks2018baselinedetectingmisclassifiedoutofdistribution}, and \textit{Mahalanobis distance}~\cite{lee2018simpleunifiedframeworkdetecting}. Performance is quantified using \textbf{AUROC}, \textbf{AUPR}, and \textbf{FPR@95\%TPR}, which together assess how well the model distinguishes low-confidence OOD predictions from confidently classified in-distribution samples. 
\end{enumerate}

This two-tier evaluation strategy enables TIE to deliver both robust OOD detection and interpretable uncertainty estimation. While the first level provides coarse anomaly rejection into the garbage class, the second level offers refined discrimination near decision boundaries.

\begin{figure*}[t]
\centering
\setlength{\tabcolsep}{1pt}
\begin{tabular}{cccc}
\includegraphics[width=\textwidth]{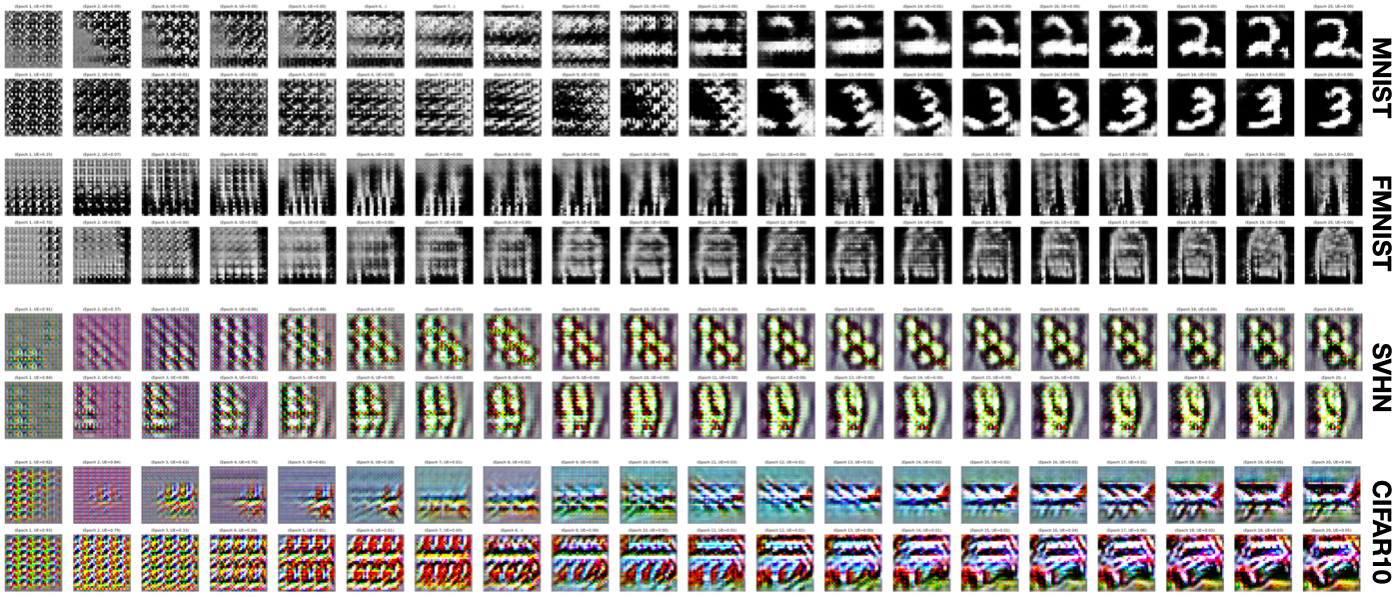} \\
\end{tabular}
\caption{
Evolution of inverted samples across epochs for select classes in the TIE framework.
}
\label{fig:qualitative_all}
\end{figure*}

\section{Qualitative Analysis of TIE}
\label{sec:qualitative_analysis}

To demonstrate the progressive refinement achieved by the proposed \textbf{TIE} framework, we present a qualitative analysis of how the reconstructed samples, associated metrics and internal representations evolve during training. 

\subsection{Visual Evolution of Inverted Samples}
\label{sec:visual_evolution}

Figure~\ref{fig:qualitative_all} shows representative reconstructions for select classes on MNIST, FashionMNIST, SVHN, and CIFAR-10 capturing how the network’s understanding of class manifolds emerges and stabilizes through training. In early epochs, the inverted samples appear noisy and unstructured, reflecting high uncertainty and weak discriminative capacity. As training progresses through the iterative \textit{TIE} cycle, the classifier and generator co-evolve: the inverted reconstructions become sharper, semantically meaningful, and increasingly representative of their true classes. 

In later epochs, while the in-distribution classes converge toward distinct, interpretable prototypes, the garbage class accumulates incoherent or ambiguous reconstructions, forming a visual boundary that isolates structured and unstructured regions in the input space. This process mirrors the classifier’s decreasing uncertainty and the progressive tightening of the dynamic exclusion threshold discussed in Section~\ref{eq:uncertainty_estimate}.

\subsection{Quantitative Evolution of Inverted Samples}
\label{sec:metric_evolution}

\begin{figure*}[t]
\centering
\includegraphics[width=\linewidth]{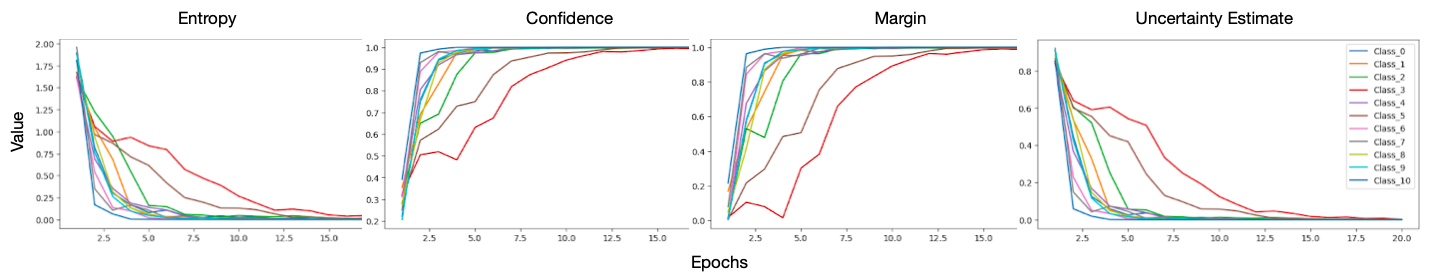}
\caption{
Evolution of averaged metrics associated with the inverted samples across epochs for all classes in MNIST. 
}
\label{fig:metric_evolution}
\end{figure*}

To complement the visual analysis, we evaluate how key predictive metrics evolve across epochs for the inverted samples of MNIST. Figure~\ref{fig:metric_evolution} presents the averaged values of \textit{entropy}, \textit{confidence}, \textit{margin}, and \textit{uncertainty estimate} for all $(n{+}1)$-classes. At the onset of training, inverted samples display high entropy and uncertainty, low confidence, and small top-1 margins—consistent with their incoherent appearance. As TIE advances, these trends systematically reverse: entropy and uncertainty decrease, while confidence and margin increase. 

This evolution indicates that the classifier’s predictions become sharper and more discriminative as the generator learns to produce increasingly class-consistent inversions.
By later epochs, all metrics stabilize, confirming convergence of the inversion process and a consistent separation between in-distribution and garbage samples. These quantitative trends align closely with the qualitative patterns observed in Figure~\ref{fig:qualitative_all}, affirming that uncertainty-guided exclusion drives both structural clarity and probabilistic calibration in reconstructed samples.

\begin{figure*}[t]
\centering
\setlength{\tabcolsep}{1pt}
\begin{tabular}{cccc}
\includegraphics[width=\textwidth]{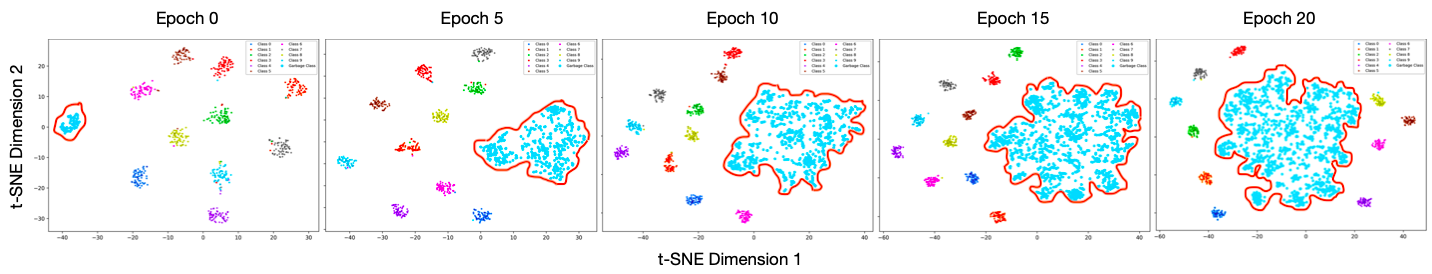} 
\end{tabular}
\caption{
t-SNE visualization of features in the latent space across epochs for all $(n+1)$ classes of MNIST shows compact in-distribution clusters surrounding the expanding garbage class region.
}
\label{fig:tsne_all}
\end{figure*}

\subsection{Feature Evolution in Latent Space}
\label{sec:tsne_evolution}

To interpret how TIE organizes the internal representations of the classifier, we visualize the latent feature embeddings using t-SNE projections across epochs.
Figure~\ref{fig:tsne_all} depicts the penultimate-layer embeddings of the classifier for all $(n{+}1)$ classes on MNIST. As the model undergoes successive \textit{Training--Inversion--Exclusion} cycles, the in-distribution classes progressively form compact and well-separated clusters, while the garbage class expands to occupy a central stage. This expansion visually reflects the exclusion mechanism, wherein uncertain or incoherent inverted samples are reassigned to the garbage class, which effectively acts as a buffer zone in latent space.

\begin{table*}[t]
\centering
\caption{
Classification accuracy (\%) for ID (diagonal) and OOD detection (off-diagonal) across all dataset pairs. Each cell includes \textbf{TIE / No TIE}, where TIE corresponds to the full Training--Inversion--Exclusion cycle, and No TIE to a baseline mode with static garbage class.
}

\label{tab:results1}
\setlength{\tabcolsep}{1.5pt}
\begin{tabular}{lcccccc}
\toprule
\textbf{Train \textbackslash\ Test} & \textbf{MNIST} & \textbf{FMNIST} & \textbf{SVHN} & \textbf{CIFAR-10} & \textbf{CIFAR-100} & \textbf{TinyImageNet-200} \\
\midrule
MNIST          & \textbf{99.1 / 97.5} & 89.5 / 23.4 & 99.1 / 31.7 & 99.4 / 36.2 & 98.6 / 28.9 & 98.3 / 25.6 \\
FMNIST         & 85.2 / 37.8 & \textbf{92.6 / 91.3} & 96.3 / 39.4 & 95.7 / 35.7 & 94.8 / 30.8 & 93.5 / 27.2 \\
SVHN           & 93.6 / 40.5 & 94.9 / 23.1 & \textbf{89.4 / 87.9} & 87.6 / 38.4 & 86.9 / 36.1 & 84.2 / 33.7 \\
CIFAR-10       & 97.8 / 46.9 & 95.7 / 36.8 & 88.2 / 25.1 & \textbf{85.5 / 84.7} & 83.6 / 39.2 & 81.4 / 37.0 \\
CIFAR-100      & 98.1 / 35.8 & 96.8 / 22.7 & 88.9 / 38.3 & 80.1 / 35.6 & \textbf{78.1 / 77.5} & 83.1 / 30.8 \\
TinyImageNet-200 & 94.2 / 32.5 & 91.9 / 28.4 & 87.4 / 37.6 & 82.9 / 36.1 & 81.7 / 24.5 & \textbf{54.8 / 52.2} \\
\bottomrule
\end{tabular}
\end{table*}

\section{Quantitative Results and Comparisons}
\label{sec:results}

We evaluate the proposed \textbf{TIE} framework on six benchmark datasets—MNIST~\citep{deng2012mnist}, FashionMNIST~\citep{xiao2017fashionmnistnovelimagedataset}, SVHN, CIFAR-10~\citep{cf}, CIFAR-100, and TinyImageNet-200—under a one-vs-rest protocol, where the model is trained using TIE Framework solely on one dataset, treated as in-distribution (ID) while the others act as out-of-distribution (OOD) sources.  

\subsection{Experimental Setup}
\label{subsec:exp_setup}

For MNIST and FashionMNIST, we employ a custom CNN comprising three convolutional and two fully connected layers, each followed by Batch Normalization and ReLU activations. For SVHN and CIFAR datasets, a CNN with five convolutional and two fully connected layers is used to accommodate higher visual complexity. While for CIFAR-100 \& TinyImageNet-200 deeper custom made CNNs with residual connections are used. The conditioned generator $\mathcal{G}_\phi$ is initialized with normally distributed weights and constructed using 4–6 transposed convolutional stages to synthesize images of the corresponding dataset resolution.  

Both the classifier and generator are optimized using Adam, with learning rates of $1\times10^{-4}$ and $1\times10^{-3}$, respectively. The slower classifier learning rate allows the generator to co-evolve and stabilize inversion dynamics. Training is performed for 20 epochs; in each epoch, approximately 1000 samples per class are added to the garbage class using a dynamic thresholding factor $\lambda=0.5$. The composite inversion loss employs weighting coefficients $\alpha$, $\beta$, and $\gamma$ of $0.1$, $1$, and $10$ for the KLD, cross-entropy, and cosine-similarity terms, respectively. $\gamma$ is progressively increased (up to $100$) to encourage wider input-space exploration.

\begin{table*}[t]
\centering
\caption{
FPR@95\%TPR (\%) across dataset pairs reported as (UE / ODIN) on the first line and (Energy / Mahalanobis) on the second line.
}
\label{tab:fpr_two_line}
\setlength{\tabcolsep}{6pt}
\renewcommand{\arraystretch}{1.6}
\resizebox{\textwidth}{!}{
\begin{tabular}{lcccccc}
\toprule
\textbf{Train \textbackslash\ Test} & \textbf{MNIST} & \textbf{FMNIST} & \textbf{SVHN} & \textbf{CIFAR-10} & \textbf{CIFAR-100} & \textbf{TinyImageNet-200} \\
\midrule
MNIST &
-- &
\begin{tabular}[c]{@{}c@{}}7.47 / 5.99 \\ 2.86 / \textbf{0.83}\end{tabular} &
\begin{tabular}[c]{@{}c@{}}45.50 / 33.75 \\ 22.47 / \textbf{0.04}\end{tabular} &
\begin{tabular}[c]{@{}c@{}}11.37 / 6.58 \\ 1.99 / \textbf{0.00}\end{tabular} &
\begin{tabular}[c]{@{}c@{}}14.30 / 9.67 \\ 3.43 / \textbf{0.00}\end{tabular} &
\begin{tabular}[c]{@{}c@{}}12.04 / 7.88 \\ 2.54 / \textbf{0.01}\end{tabular} \\

FMNIST &
\begin{tabular}[c]{@{}c@{}}66.77 / 45.65 \\ 51.21 / \textbf{8.81}\end{tabular} &
-- &
\begin{tabular}[c]{@{}c@{}}13.22 / 1.95 \\ \textbf{0.90} / 3.98\end{tabular} &
\begin{tabular}[c]{@{}c@{}}19.71 / 4.89 \\ 1.44 / \textbf{0.39}\end{tabular} &
\begin{tabular}[c]{@{}c@{}}20.53 / 5.05 \\ 1.37 / \textbf{0.52}\end{tabular} &
\begin{tabular}[c]{@{}c@{}}24.24 / 5.74 \\ 2.44 / \textbf{1.41}\end{tabular} \\

SVHN &
\begin{tabular}[c]{@{}c@{}}96.24 / 97.82 \\ 99.08 / \textbf{4.72}\end{tabular} &
\begin{tabular}[c]{@{}c@{}}\textbf{31.86} / 35.73 \\ 41.99 / 37.85\end{tabular} &
-- &
\begin{tabular}[c]{@{}c@{}}15.15 / \textbf{12.59} \\ 20.17 / 79.29\end{tabular} &
\begin{tabular}[c]{@{}c@{}}16.98 / \textbf{15.68} \\ 23.20 / 78.35\end{tabular} &
\begin{tabular}[c]{@{}c@{}}16.47 / \textbf{14.10} \\ 20.56 / 80.86\end{tabular} \\

CIFAR-10 &
\begin{tabular}[c]{@{}c@{}}77.95 / 75.18 \\ 77.60 / \textbf{12.16}\end{tabular} &
\begin{tabular}[c]{@{}c@{}}81.96 / 70.57 \\ 72.75 / \textbf{53.14}\end{tabular} &
\begin{tabular}[c]{@{}c@{}}60.79 / 63.06 \\ \textbf{57.58} / 90.56\end{tabular} &
-- &
\begin{tabular}[c]{@{}c@{}}66.32 / \textbf{66.19} \\ 67.12 / 94.45\end{tabular} &
\begin{tabular}[c]{@{}c@{}}67.74 / \textbf{66.63} \\ 67.08 / 94.76\end{tabular} \\

CIFAR-100 &
\begin{tabular}[c]{@{}c@{}}72.35 / 68.42 \\ 64.38 / \textbf{18.3}\end{tabular} &
\begin{tabular}[c]{@{}c@{}}75.91 / 70.16 \\ 73.27 / \textbf{45.8}\end{tabular} &
\begin{tabular}[c]{@{}c@{}}78.44 / 73.02 \\ \textbf{22.71} / 26.8\end{tabular} &
\begin{tabular}[c]{@{}c@{}}66.87 / \textbf{19.2} \\  64.11 / 66.2 \end{tabular} &
-- &
\begin{tabular}[c]{@{}c@{}}69.54 / \textbf{64.07} \\ 73.85 / 88.3\end{tabular} \\

TinyImageNet-200 &
\begin{tabular}[c]{@{}c@{}}71.92 / 74.65 \\ 72.44 / \textbf{21.8}\end{tabular} &
\begin{tabular}[c]{@{}c@{}}68.11 / 77.03 \\ 74.10 / \textbf{42.6}\end{tabular} &
\begin{tabular}[c]{@{}c@{}}65.47 / 60.64 \\ \textbf{25.36} / 46.3\end{tabular} &
\begin{tabular}[c]{@{}c@{}}64.20 / \textbf{20.51} \\ 63.18 / 63.5\end{tabular} &
\begin{tabular}[c]{@{}c@{}}76.88 / \textbf{63.27} \\ 71.73 / 74.6\end{tabular} &
-- \\

\bottomrule
\end{tabular}}
\end{table*}

\begin{figure*}[t]
\centering
\includegraphics[width=\linewidth]{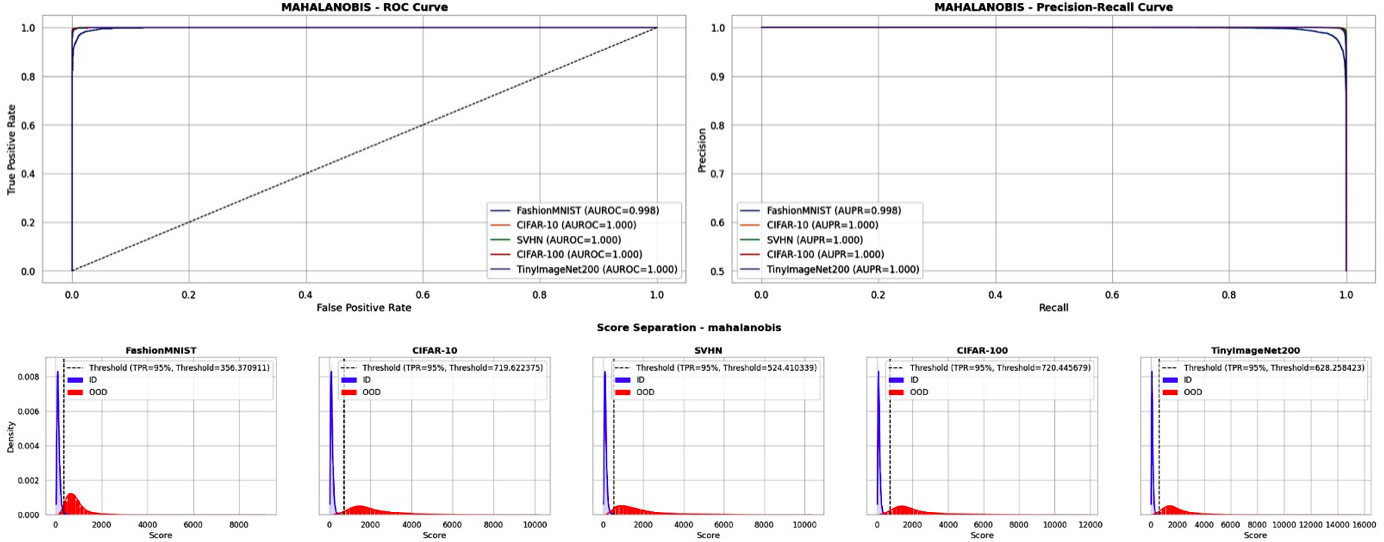}
\caption{
\textbf{Threshold-based OOD Detection on MNIST.}
Top: AUROC and AUPR curves for Mahalanobis distance across multiple OOD datasets.
Bottom: Score separation plots showing distribution of Mahalanobis distances for in-distribution (blue) and OOD (red) samples.
}
\label{fig:mnist_mahalanobis}
\end{figure*}

\begin{figure*}[t]
\centering
\includegraphics[width=\linewidth]{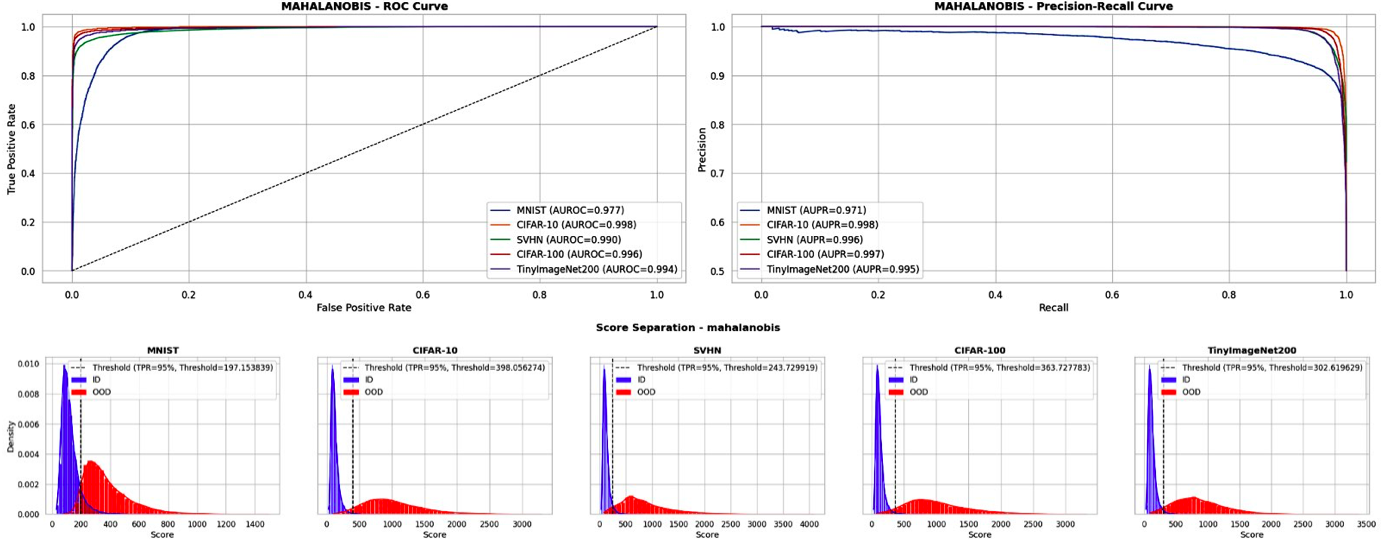}
\caption{
\textbf{Threshold-based OOD Detection on FashionMNIST.}
Top: AUROC and AUPR curves for Mahalanobis distance across multiple OOD datasets.
Bottom: Score separation plots illustrating clear distinction between in-distribution (blue) and OOD (red) samples. 
}
\label{fig:fmnist_mahalanobis}
\end{figure*}

\subsection{OOD Detection into Garbage Class}
\label{subsec:direct_ood}

At the first evaluation level, we assess the model’s ability to directly classify out-of-distribution inputs into the garbage class while maintaining strong in-distribution accuracy.  
Table~\ref{tab:results1} reports both in-distribution (diagonal) and cross-domain OOD (off-diagonal) accuracies for all dataset pairs.  
High diagonal scores indicate that TIE retains excellent ID performance, while strong off-diagonal accuracies confirm its capacity to isolate anomalous inputs. 
These results validate the joint learning of the classifier and generator, which allows the system to simultaneously preserve discriminative structure and enforce exclusion.

Overall, TIE demonstrates excellent in-distribution classification accuracy on simpler datasets such as MNIST and FashionMNIST, achieving over 99\% and 92\% respectively. However, cross-domain evaluation reveals interesting trends: MNIST performs slightly worse on FashionMNIST owing to their shared grayscale characteristics. Conversely, SVHN struggles when evaluated against MNIST, as both represent digit domains with differing statistical structures. Since all models are trained from scratch without pretrained backbones, the in-distribution performance on complex datasets such as CIFAR-100 and TinyImageNet remains lower, yet even these models exhibit robust OOD discrimination when evaluated against other datasets. 

\textbf{Baseline Comparisons:} We compare TIE against a baseline $(n{+}1)$-class classifier trained with a static garbage class. Removing TIE results in a dramatic collapse in OOD recognition accuracy by \textbf{50–70\%} as shown in Table~\ref{tab:results1} with minimal impact on in-distribution performance. TIE’s dynamic refinement cycle, powered by inversion and adaptive thresholding, proves essential for carving out meaningful decision boundaries across both ID and OOD regions.

\subsection{Threshold-Based OOD Detection}
\label{subsec:threshold_ood}

While most OOD samples are directly excluded into the garbage class during inference, a subset of ambiguous inputs may still be misclassified into in-distribution categories with relatively low confidence and high uncertainty. This evaluation level addresses such borderline cases to assess TIE’s ability to perform fine-grained OOD discrimination. For this analysis, we discard the $(n{+}1)$\textsuperscript{th} garbage class logits and normalize the remaining $n$ outputs before performing threshold-based OOD detection. 

\textbf{Metric Comparisons:}
In Table \ref{tab:fpr_two_line}, we further compare the performance of multiple OOD detection metrics—\textit{UE}, \textit{ODIN}, \textit{Energy}, and \textit{Mahalanobis Distance}—in terms of FPR@95\%TPR across all dataset pairs to quantify TIE’s ability to distinguish confidently classified in-distribution samples from uncertain, misclassified OOD instances. Since this evaluation is performed only on the small subset of OOD samples misclassified into in-distribution classes, the resulting AUROC and AUPR values are expected to be lower, and FPR@95\%TPR to be higher compared to standard full-dataset OOD benchmarks.

Across both MNIST and FashionMNIST evaluations, Mahalanobis-based thresholding yields outstanding fine-grained OOD discrimination as shown in Figures~\ref{fig:mnist_mahalanobis}–\ref{fig:fmnist_mahalanobis}.
For MNIST, AUROC and AUPR values are 0.998 for FashionMNIST and 1 for all other datasets, with corresponding average FPR@95\%TPR close to zero(0.00176). While FashionMNIST achieves average AUROC, AUPR and FPR@95\%TPR scores of 0.9911, 0.9916 and 0.0302 respectively across all other datasets. The lower panels in Figures~\ref{fig:mnist_mahalanobis}–\ref{fig:fmnist_mahalanobis} reveal distinct score density gaps, where the Mahalanobis distance for ID samples remains tightly clustered while OOD scores spread widely across higher values. This pronounced margin confirms that the feature embeddings learned under TIE are highly structured and uncertainty-aware, providing an intrinsic mechanism for OOD discrimination without explicit calibration. Across complex datasets such as CIFAR-10, CIFAR-100, and TinyImageNet-200, ODIN achieves lower FPR@95\%TPR, indicating better separation of near-boundary OOD samples. In contrast, when SVHN acts as the OOD dataset, the Energy score yields the lowest FPR values, suggesting that energy-based scoring is particularly effective for detecting digit-domain anomalies.

\section{Conclusion and Future Work}
\label{sec:conclusion}

In this paper, we introduced \textbf{TIE} as a unified framework for visually interpretable and uncertainty-aware out-of-distribution detection by embedding inversion into the classifier training process, coupled with dynamic uncertainty-based exclusion.

Future work can decompose the global garbage class into $n+n$ separate garbage classes—one corresponding to each in-distribution category—for fine-grained OOD detection, reducing dependence on threshold-based post-processing.

\bibliography{iclr2026_conference}
\bibliographystyle{iclr2026_conference}
\newpage
\appendix

\end{document}